\documentclass[10pt]{article} 
\usepackage[accepted]{rlc}

\usepackage{amssymb}            
\usepackage{mathtools}          
\usepackage{mathrsfs}           
\mathtoolsset{showonlyrefs}     
\usepackage{graphicx}           
\usepackage{subcaption}         
\usepackage[space]{grffile}     
\usepackage{url}                
 
\usepackage{todonotes}
\usepackage{xurl} 
\usepackage{booktabs}       
\usepackage{amsfonts}       
\usepackage{nicefrac}       
\usepackage{microtype}      
\usepackage{xcolor}         
\usepackage{amsmath,amsthm,amssymb}
\usepackage{dsfont}
\usepackage{caption}
\usepackage{adjustbox}
\usepackage{enumitem}
\usepackage{hyperref}
\usepackage{colortbl}
\usepackage{highlight}
\usepackage{float}
\usepackage{longtable}

\usepackage{wrapfig}

\usepackage{multirow}
\usepackage{array}
\newcolumntype{y}[1]{>{\centering\let\newline\\\arraybackslash\hspace{0pt}}p{#1}}
\renewcommand*{\arraystretch}{1.2}

\newcommand\blfootnote[1]{%
  \begingroup
  \begin{NoHyper}
  \renewcommand\thefootnote{}\footnote{#1}%
  \addtocounter{footnote}{-1}%
  \end{NoHyper}
  \endgroup
}

\newcommand{\IIWA}{\texttt{IIWA}}
\newcommand{\Jaco}{\texttt{Jaco}}
\newcommand{\Panda}{\texttt{Panda}}
\newcommand{\Kinova}{\texttt{Gen3}}

\newcommand{\boxObject}{\texttt{box}}
\newcommand{\hollowBox}{\texttt{hollow\_box}}
\newcommand{\dumbbell}{\texttt{dumbbell}}
\newcommand{\plate}{\texttt{plate}}

\newcommand{\objectWall}{\texttt{object\_wall}}
\newcommand{\objectDoor}{\texttt{object\_door}}
\newcommand{\goalWall}{\texttt{goal\_wall}}
\newcommand{\noObstacle}{\texttt{no\_obstacle}}

\newcommand{\pickPlace}{\texttt{pick\_and\_place}}

\newcommand{\push}{\texttt{push}}
\newcommand{\shelf}{\texttt{shelf}}
\newcommand{\trashcan}{\texttt{trash\_can}}

\title{Robotic Manipulation Datasets for\\Offline Compositional Reinforcement Learning}

\author{
  $\!\!$Marcel Hussing{\normalfont $^\dagger$,}\\
  University of Pennsylvania \\ 
  \texttt{mhussing@seas.upenn.edu} \\
  \And
  Jorge Mendez-Mendez{\normalfont $^\dagger$,}\\
  Massachusetts Institute of Technology \\
  \texttt{jmendezm@mit.edu} \\
  \And
  Anisha Singrodia\\
  University of Pennsylvania \\
  \texttt{singroa@seas.upenn.edu} \\
  \And
  Cassandra Kent\\
  University of Pennsylvania \\
  \texttt{dekent@seas.upenn.edu} 
  \And
  Eric Eaton \\
  University of Pennsylvania \\
  \texttt{eeaton@seas.upenn.edu} 
}

\begin{document}
\maketitle

\vspace{-.5em}
\begin{abstract}

\protect\blfootnote{\hspace{-0.4em}$\dagger$ The two first authors contributed equally to this work.} \hspace{-0.55em}
Offline reinforcement learning (RL) is a promising direction that allows RL agents to pre-train on large datasets, avoiding the recurrence of expensive data collection. To advance the field, it is crucial to generate large-scale datasets. Compositional RL is particularly appealing for generating such large datasets, since 1)~it permits creating many tasks from few components, 2)~the task structure may enable trained agents to solve new tasks by combining relevant learned components, and 3)~the compositional dimensions provide a notion of task relatedness. This paper provides four offline RL datasets for simulated robotic manipulation created using the $256$ tasks from CompoSuite~\citep{mendez2022composuite}. Each dataset is collected from an agent with a different degree of performance, and consists of $256$ million transitions. We provide training and evaluation settings for assessing an agent's ability to learn compositional task policies. Our benchmarking experiments show that current offline RL methods can learn the training tasks to some extent and that compositional methods outperform non-compositional methods. Yet current methods are unable to extract the compositional structure to generalize to unseen tasks, highlighting a need for future research in offline compositional RL. 
\end{abstract}


\section{Introduction} \label{sec:intro}

Large-scale data has generated much of the success of deep learning. 
We would expect robot learning techniques to similarly leverage vast amounts of data to solve multitudes of real-world problems. However, 
generating datasets for robotics is expensive and time consuming, even in simulation. Large-scale data collection is imperative to maximizing the utility of deep learning for robotics.

Much of the efforts in deep learning research for robotics have been devoted to reinforcement learning (RL).  
However, online RL methods require the agent to collect data over time, and therefore each new online RL experiment requires a new round of large-scale data collection. Offline RL approaches~\citep{LangeGabelRiedmiller2011chapter, fujimoto2019off} train on a fixed (previously collected) dataset, potentially permitting to learn high-quality policies without the need to obtain additional data. Once an agent has been pre-trained on offline data, its model can be fine-tuned to unseen tasks in the real world with little additional data~\citep{chebotar2021actionable}. Despite these advantages, the offline setting comes with its own challenges. First, offline RL requires large datasets~\citep{fu2020d4rl} labeled with reward functions. Image labels, the computer vision counterpart, can easily be crowdsourced, which has facilitated the creation of large vision datasets; crowdsourcing is not readily applicable to RL rewards. 
Second, offline RL agents are not allowed to explore new states during training, and must generalize to unseen states at evaluation time. 
Notably, unlike in supervised settings, RL agents do not make a single prediction on a new state, but instead the actions they choose lead them to traverse the state space, moving them increasingly far away from the original training distribution.
This leads to a unique form of distribution shift. These issues are exacerbated by the fact that standard offline RL evaluations are limited to single-task problems,
further restricting the scale of current datasets.

To address these issues, we consider compositional agents and environments. A \textit{compositional agent} decomposes complex problems into components, re-composes the components to solve the problems, and re-uses the acquired knowledge throughout the state space, improving state generalization. Further, compositional RL agents exhibit sample efficiency improvements in multi-task and lifelong RL via generalizable components and behaviors that can be combined to solve new tasks~\citep{mendez2022modular}. 
On the other hand, \textit{compositional  environments} offer re-usability of reward functions to induce a plethora of training behaviors~\citep{mendez2022composuite}. They also enable the creation of numerous tasks with a clear notion of task relatedness along the different compositional dimensions, which is useful for selective transfer and for analyzing performance.

To facilitate the combined study of offline RL and compositionality, we provide multiple datasets collected using CompoSuite~\citep{mendez2022composuite}---a simulated robotic manipulation  benchmark designed for studying online compositional RL---and experiment scenarios designed to answer questions related to the interplay of the two fields. Specifically, we contribute the following\footnote{Datasets are available at \href{https://datadryad.org/stash/dataset/doi:10.5061/dryad.9cnp5hqps}{datadryad.org/stash/dataset/doi:10.5061/dryad.9cnp5hqps}; train-test split lists and code for the experiments can be found at \href{https://github.com/lifelong-ml/offline-compositional-rl-datasets}{github.com/lifelong-ml/offline-compositional-rl-datasets}.
}:

\begin{enumerate}[leftmargin=*,noitemsep,topsep=0pt,parsep=0pt,partopsep=0pt]
    \item Four datasets of varying performance with trajectories from each of the $256$ CompoSuite tasks,
    \item Training-test split lists for evaluation to ensure comparability and reproducibility of results, and 
    \item An evaluation demonstrating the utility of our datasets for offline compositional RL research, and the (relatively) poor ability of current offline RL techniques to leverage compositional structures. These results validate both the learnability and difficulty of the datasets using common learning techniques, and demonstrate the need for improved algorithms for offline compositional RL.
\end{enumerate}


\section{Preliminaries} 
\label{sec:preliminaries}

\begin{figure}
\captionsetup[subfigure]{aboveskip=1pt}
    \begin{subfigure}[b]{0.25\textwidth}
        \centering
            \captionsetup{width=0.95\linewidth,justification=centering,singlelinecheck=false}
            \includegraphics[width=0.95\linewidth, trim={0cm 5cm 0cm 4cm}, clip]{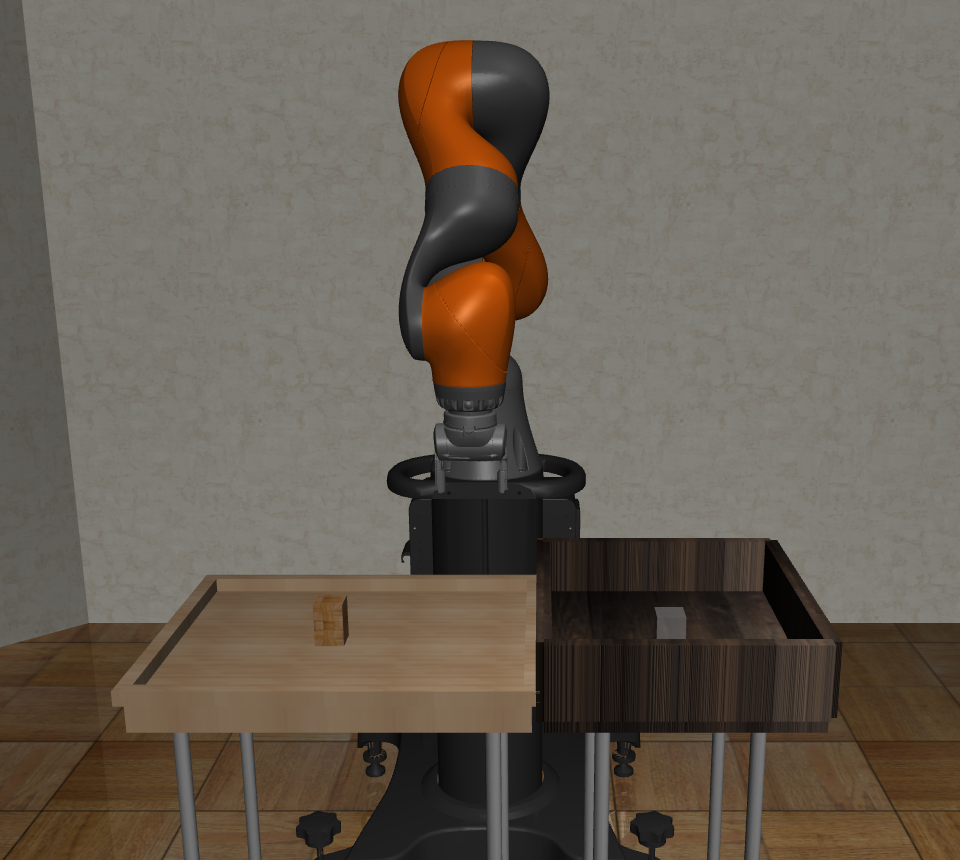}
            \caption*{{\scriptsize$\langle$\IIWA{}, \boxObject{}, \noObstacle{}, \pickPlace{}$\rangle$}}
            \label{fig:arena1}
    \end{subfigure}%
    \begin{subfigure}[b]{0.25\textwidth}
        \centering
            \captionsetup{width=0.95\linewidth,justification=centering,singlelinecheck=false}
            \includegraphics[width=0.95\linewidth, trim={0cm 5cm 0cm 4cm}, clip]{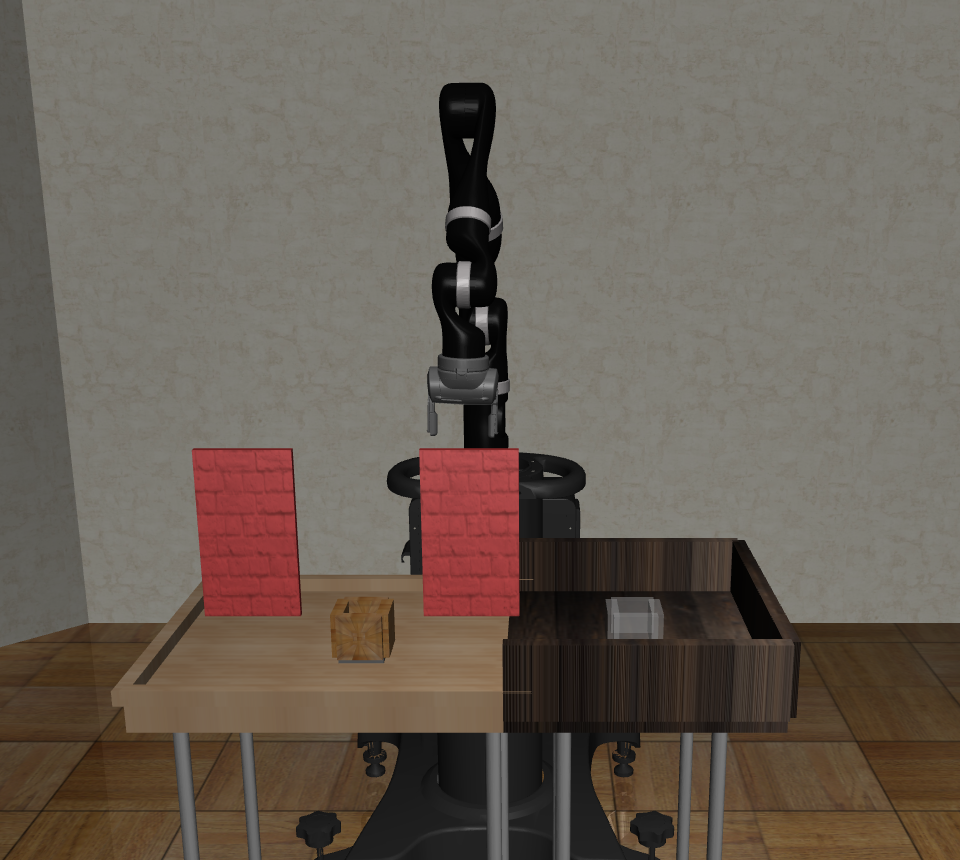}
            \caption*{{\scriptsize$\langle$\Jaco{}, \hollowBox{}, \objectDoor{}, \push{}$\rangle$}}
    \end{subfigure}%
    \begin{subfigure}[b]{0.25\textwidth}
        \centering
            \captionsetup{width=0.95\linewidth,justification=centering,singlelinecheck=false}
            \includegraphics[width=0.95\linewidth, trim={0cm 5cm 0cm 4cm}, clip]{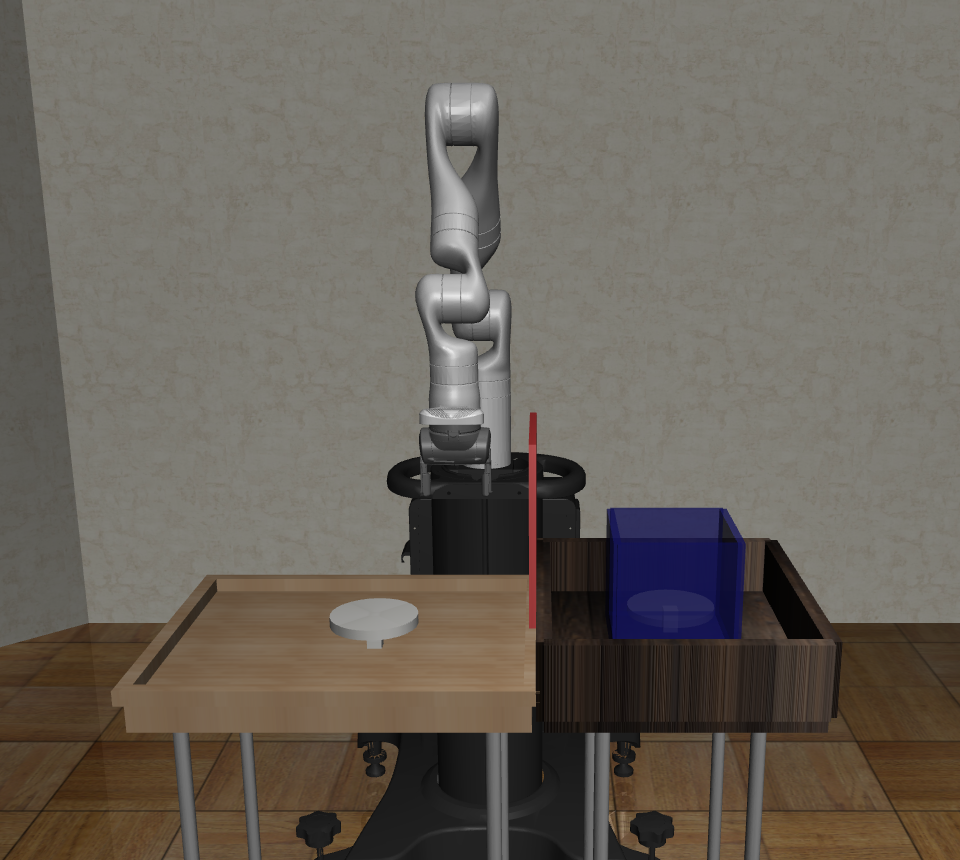}
            \caption*{{\scriptsize$\langle$\Kinova{}, \plate{}, \goalWall{}, \trashcan{}$\rangle$}}
    \end{subfigure}%
    \begin{subfigure}[b]{0.25\textwidth}
        \centering
            \captionsetup{width=0.95\linewidth,justification=centering,singlelinecheck=false}
            \includegraphics[width=0.95\linewidth, trim={0cm 5cm 0cm 4cm}, clip]{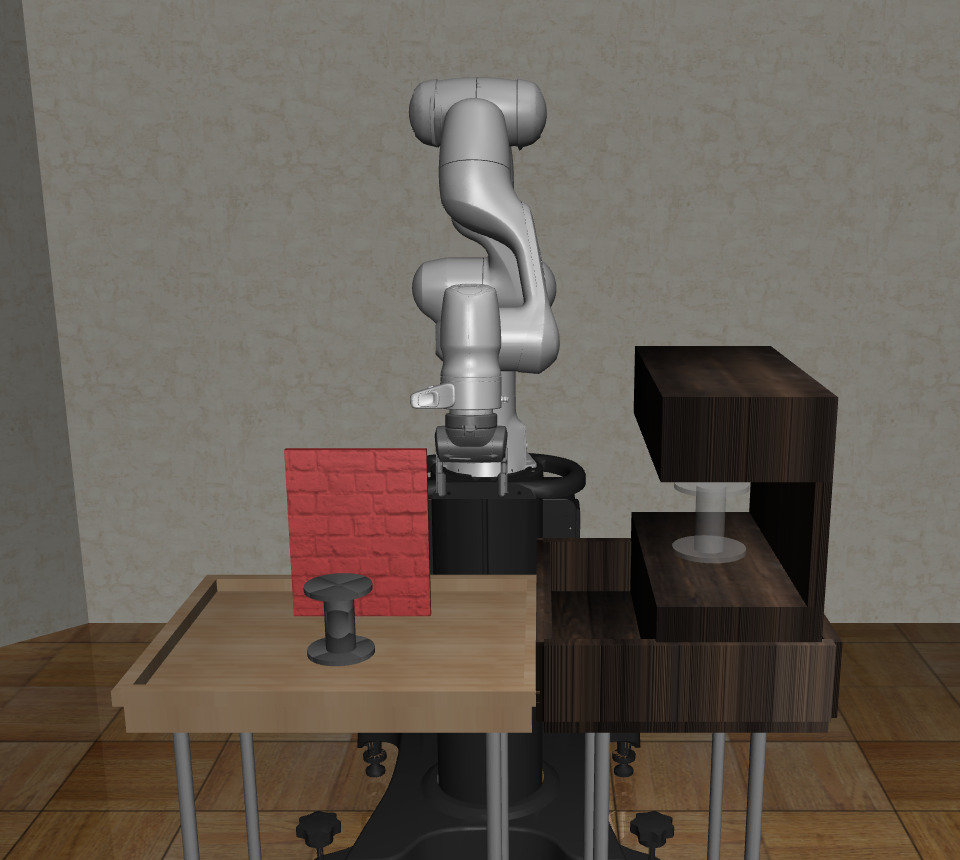}
            \caption*{{\scriptsize$\langle$\Panda{}, \dumbbell{}, \objectWall{}, \shelf{}$\rangle$}}
    \end{subfigure}
    \caption{Examples of four CompoSuite tasks, showing each task's initial state. Each task is composed of one element from each of four compositional axes, involving a {\em robot} (\IIWA{}, \Jaco{}, \Kinova{}, or \Panda{}) manipulating an {\em object} (\boxObject{}, \hollowBox{}, \plate{}, or \dumbbell{}) while avoiding an {\em obstacle} (\noObstacle{}, \objectDoor{}, \goalWall{}, \objectWall{}) to achieve a specific {\em objective} (\pickPlace{}, \push{}, \trashcan{}, or \shelf{}). Images from \citet{mendez2022composuite}.}\label{fig:arenas}
\end{figure}

\textbf{Offline RL}~~Standard \textit{online} RL solves a Markov decision process ${\mathcal{M}=\{\mathcal{S}, \mathcal{A}, R, \mathcal{P}, \gamma, \mu\}}$ via direct interaction with $\mathcal{M}$, where $\mathcal{S}$ is the state space, $\mathcal{A}$ is the action space, $R$ is the reward function, $\mathcal{P}$ are the transition probabilities, $\gamma$ is a  discount factor, and $\mu$ is the distribution over starting states. In \textit{offline} RL, the agent does not have access to $\mathcal{M}$ for training, but instead receives a dataset $\mathcal{D}=\{(s_i,a_i,s'_i,r_i)\}_{i=1}^{N}$ of transition tuples, where $s_i'$ and $r_i$ are the state and reward obtained by executing action $a_i$ in state $s_i$ using an unknown behavioral policy $a_i\sim\pi_{\beta}(s_i)$ in $\mathcal{M}$. Consequently, $\mathcal{D}$ is a sample from the distribution $d^{\pi_{\beta}}(s)\pi_{\beta}(s, a)$, where $d^{\pi_{\beta}}(s)$ is the marginal state distribution induced by $\pi_{\beta}$. The goal in offline RL is to find an optimal policy $\pi^{*}$ which maximizes the expected cumulative return $J_{\pi} = \mathbb{E}[\sum_{t=0}^{\infty} \gamma^t r(s_t, a_t)]$ in $\mathcal{M}$ \emph{without interacting with $\mathcal{M}$}. 

\textbf{Functional composition in RL}~~Unlike traditional temporal sequencing of skills, functionally compositional RL considers the composition of functional transformations of the state by a chain of computations that results in a chosen action~\citep{mendez2022modular}. These functional modules are akin to functions in programming, which consume as inputs the outputs of other functions and produce inputs to yet other functions. At each timestep, multiple functions are involved in computing the action to take from the current state; compare this to temporal composition (e.g., in the Options framework \citep{sutton1999options, bacon2017option}) in which only one module (an option) is active at each time. A set of RL tasks related via functional composition can be described formally as a compositional problem graph whose paths represent the transformations required to solve each task. 

\textbf{CompoSuite benchmark for compositional RL}~~ CompoSuite~\citep{mendez2022composuite} is a recent simulated robotics benchmark for RL built on top of robosuite~\citep{yuke2020robosuite}, designed to study functional composition in RL. Every CompoSuite task is created by composing elements of four different axes: a \texttt{robot} manipulator that moves an \texttt{object} to achieve an \texttt{objective} while avoiding an \texttt{obstacle}. Each axis consists of four elements, for a total of $256$ tasks (Figure~\ref{fig:arenas}). For a given \texttt{objective}, the reward function is constant across other axes, making it easy to scale the number of tasks without the need to craft labeling functions for each individual task.  

\section{Datasets and experimental setup for offline compositional RL}

\begin{figure}[t]
    \centering
    \includegraphics[width=\textwidth,clip,trim=2cm 3.8cm 2cm 4cm]{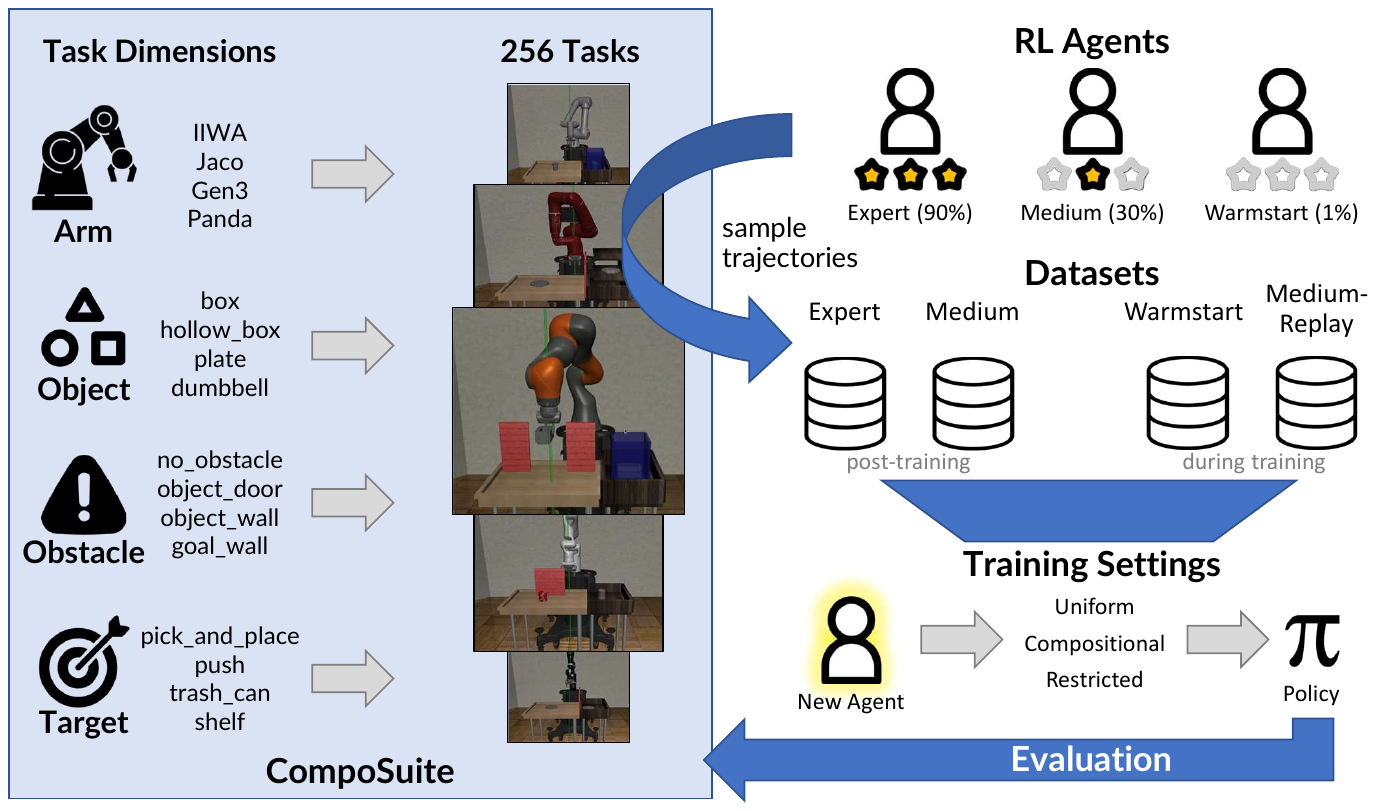} 
    \caption{An overview of our dataset creation and training process. Manipulation tasks vary along four compositional dimensions, as taken from CompoSuite. Trajectories are sampled from pre-trained PPO agents, forming four different datasets of varying difficulties (Section~\protect\ref{sec:datasetdesc}). Three different training settings (Section~\protect\ref{sec:trainlists}) provide different views into these data for training, and evaluation of the learned policies is performed on the CompoSuite simulator using mujoco. }
    \label{fig:Overview}
\end{figure}

We elaborate on the specific training setting we consider and structure of the datasets we provide, and detail several reproducible experiment configurations for analyzing offline compositional RL. Figure~\ref{fig:Overview} provides an overview of the dataset collection process and its use in training and evaluation.  

\subsection{Data shape and spaces}

Following \citet{fu2020d4rl}, we collect one million transitions for each task, totaling $256$ million transitions per dataset. Every transition contains: observation, action, reward, next observation, timeout indicator, and terminal indicator. Observations are vectors of size $93$ containing proprioceptive robot information such as joint and finger positions and velocities, absolute and relative object, obstacle, and goal positions, and a multi-hot task indicator to identify the elements of the current task. The action space is eight-dimensional; the first seven dimensions correspond to target joint angles of the 7-DoF robots for joint position control, and the last dimension is a gripper action. Tasks use dense rewards to ensure that every transition has a non-zero reward. Rewards are specific to each objective and encourage the learning of a policy in stages (e.g., the rewards for pick-and-place encourage first reaching the object, then grasping the object, lifting the object, and finally approaching the target).
The reward for being in a goal-satisfying state is $1$ for all tasks. Episodes time out after $500$ timesteps, and \texttt{push} tasks additionally terminate if the grasped object is lifted from the table.

\subsection{Data collection}
\label{sec:datasetdesc}

To collect our datasets, we trained agents using standard online RL methods to obtain the behavioral policies $\pi_{\beta}$. For three of the datasets, we trained a single agent via proximal policy optimization (PPO; \citealp{schulman2017proximal}) across all tasks, storing trained policies at various levels of performance for each task. PPO can be parallelized and offers a fast algorithm to obtain these policies in terms of wall-clock time. To ensure that the agent would achieve high success rate on all tasks, we used the compositional neural network architecture of~\citet{mendez2022modular}. For the fourth dataset, we trained a separate agent on each task via soft actor critic (SAC; \citealp{haarnoja2018sac}) to simulate a warmstart scenario in which some data from an RL run is available. Here, we use SAC due to its sample efficiency. Concretely, we provide the following four datasets.

\begin{itemize}[noitemsep,topsep=0pt,parsep=0pt,partopsep=0pt,leftmargin=*]
    \item \textbf{Expert dataset:} Transitions from an agent trained to achieve $90\%$ success on every task. 
    \item \textbf{Medium dataset:} Transitions from an agent trained to achieve $30\%$ success on every task.
    \item \textbf{Warmstart dataset:} Transitions that were stored during the training of one SAC agent per task for 1 million steps. The average success rate across all trajectories is in the order of 1\%.
    \item \textbf{Medium-replay (subsampled) dataset:} Transitions that were stored during the training of an agent up to $30\%$ success. For tasks that required more than one million steps to achieve $30\%$ success, the one million transitions were obtained by uniformly sampling trajectories.
\end{itemize}

The different datasets were collected to serve various research purposes. In the real world, expert data is rarely available. Instead, datasets have varying levels of performance, represented by the  expert, medium, and warmstart datasets. This allows for the construction of training sets containing data from trajectories of varying success rates and task progress (as discussed in Section~\ref{sec:trainlists}), which both offers realistic data collection settings and lets researchers experiment with diverse levels of difficulty of offline RL tasks. We chose to replace the random agents commonly used in the literature (e.g., \citealp{fu2020d4rl}) with warmstart agents trained for a short period to ensure that the data contained diverse states covering the various stages of each task---because the tasks are long-horizon, a random policy would only cover a small portion of the state space, since it would never even grasp the object. Warmstart agents simulate a setting where a researcher can do \emph{some} online-RL but is limited in time.

In addition, the medium-replay dataset contains data that an online RL agent would see during training, exhibiting varied levels of proficiency at solving the task. Intuitively, this should be sufficient to learn good policies via offline RL, yet current approaches struggle in this setting~\citep{fujimoto2019off,fu2020d4rl}. Note that since our agents require substantially more than one million samples to converge in many of the tasks, the medium-replay dataset subsamples trajectories observed during training. Since some trajectories from \push{} tasks might be truncated (due to the termination condition), we artificially truncate the last subsampled trajectory and place a time-out at the final position. This might, in rare cases, lead to one incorrect trajectory if the datasets are used for explicitly finite horizon RL experimentation. However, the truncation ensures a consistent dataset size of one million across tasks and compatibility with other standard code implementations.

\subsection{Training task lists and multi-task training} \label{sec:trainlists}

We consider multiple training settings to analyze an agent's ability to functionally decompose a task and re-use its acquired knowledge. These settings are represented by different samplings of tasks across the various datasets. To facilitate comparability of results, we provide lists splitting the tasks into training and zero-shot tasks, analogous to train-test splits in supervised learning problems.
Any of the following sampling techniques can be used with any of the datasets from section~\ref{sec:datasetdesc}.  

\textbf{Uniform sampling}~~ This standard multi-task setting is used to evaluate training performance and zero-shot generalization. Akin to data splits in supervised learning, the agent trains offline on $224$ tasks and is evaluated for generalization to $32$ online test tasks without any data for those tasks. 

\textbf{Compositional sampling}~~ A more realistic setting should not assume access to data of equal performance for every task. To simulate this scenario, we split the data into $76$ training tasks from the expert dataset, $148$ additional training tasks from one of the other (non-expert) datasets, and $32$ zero-shot tasks. The $76$ expert tasks contain all $16$ CompoSuite components in equal proportions. This setting acts as a proxy for measuring compositionality in a learning approach; a model that can successfully decompose its knowledge about successful executions from the expert tasks into the relevant components should be able to combine this knowledge with the noisier information from remaining tasks to compositionally generalize to those and the unseen tasks. Note that, if non-expert tasks are drawn from the warmstart dataset, the combined dataset has similar average success as the medium dataset but with a substantially different success distribution across tasks. 

\textbf{Restricted sampling}~~ Similar to the equivalent setting in online CompoSuite, restricted sampling constitutes a harder setting to evaluate an agent's ability to extract compositional information. This is achieved by restricting the training dataset to be smaller and to contain only a single task for a specific element. For example, if the selected element is the \IIWA{} arm, then the training set contains exactly one task which uses an \IIWA{} arm and $55$ tasks that use other arms. The training set contains a total of $56$ tasks while the zero-shot set contains the remaining $63$ tasks that contain the \IIWA{} arm. 


\section{Experiments} \label{sec:experiments}
\label{sec:result}

\subsection{Implementation and experiment details} \label{sec:impl}

We evaluated various settings from Section~\ref{sec:trainlists} over three different random seeds controlling the choice of train-test split list, network parameter initialization, and data sampling within each algorithm---because our comparisons are over  large numbers of tasks, the results have low variance and so three seeds are sufficient to show general trends. We consider four baselines\footnote{We additionally ran the evaluations using conservative Q-learning~\citep{kumar2020cql}, but found that it attained 0\% success on all settings (including the training tasks with Expert data), so we omit it from our results.}: Behavioral Cloning (BC), Compositional BC (CP-BC), Implicit Q-Learning (IQL; \citealp{kostrikov2022iql}) and Compositional IQL (CP-IQL), using the d3rlpy implementations~\citep{seno2021d3rlpy} (hyperparameters in Appendix~\ref{sec:hyperparameters}). BC imitates the behavioral policy $\pi_{\beta}$ by learning to predict the correct action given a state from the dataset, and we expect it to perform well given high-performance data.  IQL is an offline RL baseline designed to generalize beyond the training data distribution and is expected to achieve better performance given non-expert data. These two baselines use standard multilayer perceptrons (MLP) to encode policies and value functions. The CP versions of the algorithms instead employ a compositional neural network architecture as described by~\citet{mendez2022modular, mendez2022composuite} for all networks (see Appendix~\ref{sec:DetailsCompositionalPolicy} for details). The compositional network architecture consists of hierarchically stacked modules that correspond to the various elements in CompoSuite. Each module operates in two stages: the pre-processing stage is an MLP that takes as input the module-specific state (e.g., object modules take only the object state as input); the post-processing stage is a second MLP that takes as input the concatenation of the output of the previous module (in the hierarchy) and the output of the pre-processing stage. Intuitively, encoding the tasks' inherent compositional structures into the policies should facilitate transfer to unseen tasks.

We trained each agent simultaneously on a subset of the $256$ tasks, and evaluated it on held-out tasks per the task lists from Section~\ref{sec:trainlists}. All BC and IQL agents were trained for  50,000 and 300,000 update steps respectively using a batch size of $\# \text{training tasks} \times 256$. 
Trained agents are evaluated online using CompoSuite~\citep{mendez2022composuite}, with the metrics from Appendix~\ref{sec:metrics}. We report mean cumulative return and success rate over one evaluation trajectory per task for train and test tasks. 

\subsection{Experimental results}

\begin{table}[t!]
    \caption{Test and training return and success rates achieved by Behavioral Cloning (BC), Implicit Q-Learning (IQL), Compositional BC (CP-BC), and Compositional IQL (CP-IQL) on the various datasets using 224 training tasks and 32 test tasks. All agents achieve decent success and generalize when given access to expert data (sub-table $1$). IQL agents strictly outperform BC on the medium and replay datasets (sub-tables $2$ and $3$). When having to extract compositional information from expert data, the compositional policy yields some benefits over feed-forward networks but is still far from optimal (sub-table 4). Success rates are shaded from green (100\%) to yellow (50\%) to red (0\%). All values represent mean $\pm$ standard deviation.\\}
    \label{tab:defaultres}
    \centering
        \begin{tabular}{y{110pt}|y{59pt}y{59pt}y{59pt}y{59pt}}
         & \multicolumn{4}{c}{Dataset: Expert; Sampling: Uniform}\\
         & Train Return & Test Return & Train Success & Test Success \\
         \hline
        Behavioral Cloning & $339.05 $ {\tiny $\pm4.26$} & $297.29  $ {\tiny $\pm7.18\,\,$} & \shadecell{0.87} {\tiny $\pm0.01$} & \shadecell{0.73} {\tiny $\pm0.02$} \\
        Implicit Q-Learning & $264.97 $ {\tiny $\pm2.16$} & $279.67  $ {\tiny $\pm33.92$} & \shadecell{0.65} {\tiny $\pm0.01$} & \shadecell{0.68} {\tiny $\pm0.07$} \\
        CP Behavioral Cloning & $380.42 $ {\tiny $\pm2.44$} & $354.61  $ {\tiny $\pm11.23$} & \shadecell{0.97} {\tiny $\pm0.01$} & \shadecell{0.88} {\tiny $\pm0.05$} \\
        CP Implicit Q-Learning & $351.62 $ {\tiny $\pm1.98$} & $345.19$ {\tiny $\pm10.16$} & \shadecell{0.90} {\tiny $\pm0.01$} & \shadecell{0.86} {\tiny $\pm0.03$} \\
        \end{tabular} \\
        \vspace{0.5em}
        \begin{tabular}{y{110pt}|y{59pt}y{59pt}y{59pt}y{59pt}}
         & \multicolumn{4}{c}{Dataset: Medium; Sampling: Uniform}\\
         & Train Return & Test Return & Train Success & Test Success \\
         \hline
        Behavioral Cloning & $190.84 $ {\tiny $\pm8.40\,\,\, $} & $162.93  $ {\tiny $\pm6.11\,\, $} & \shadecell{0.24} {\tiny $\pm0.03$} & \shadecell{0.21} {\tiny $\pm0.06$} \\
        Implicit Q-Learning & $176.84 $ {\tiny $\pm11.46$} & $150.66  $ {\tiny $\pm22.31$} & \shadecell{0.30} {\tiny $\pm0.02$} & \shadecell{0.24} {\tiny $\pm0.02$} \\
        CP Behavioral Cloning & $211.10 $ {\tiny $\pm3.27$} & $190.04 $ {\tiny $\pm21.61$} & \shadecell{0.28} {\tiny $\pm0.02$} & \shadecell{0.22} {\tiny $\pm0.08$} \\
        CP Implicit Q-Learning & $223.44 $ {\tiny $\pm4.23$} & $196.98$ {\tiny $\pm30.63$} & \shadecell{0.47} {\tiny $\pm0.02$} & \shadecell{0.38} {\tiny $\pm0.1$} \\
        \end{tabular} \\
        \vspace{0.5em}
        \begin{tabular}{y{110pt}|y{59pt}y{59pt}y{59pt}y{59pt}}
         & \multicolumn{4}{c}{Dataset: Medium-Replay; Sampling: Uniform}\\
         & Train Return & Test Return & Train Success & Test Success \\
         \hline
        Behavioral Cloning & $102.65 $ {\tiny $\pm4.63$} & \,\,$95.04  $ {\tiny $\pm12.00$} & \shadecell{0.00} {\tiny $\pm0.01$} & \shadecell{0.00} {\tiny $\pm  0.00$} \\
        Implicit Q-Learning & $138.37 $ {\tiny $\pm1.68$} & $142.35  $ {\tiny $\pm23.51$} & \shadecell{0.10} {\tiny $\pm0.02$} & \shadecell{0.09} {\tiny $\pm0.05$} \\
        CP Behavioral Cloning & $95.31 $ {\tiny $\pm1.04$} & \,\,$91.60  $ {\tiny $\pm5.02$} & \shadecell{0.00} {\tiny $\pm0.00$} & \shadecell{0.00} {\tiny $\pm0.00$} \\
        CP Implicit Q-Learning & $102.66 $ {\tiny $\pm10.11$} & \,\,$99.08  $ {\tiny $\pm20.67$} & \shadecell{0.09} {\tiny $\pm0.03$} & \shadecell{0.09} {\tiny $\pm0.03$} \\
        \end{tabular}\\
        \vspace{0.5em}
        \begin{tabular}{y{110pt}|y{59pt}y{59pt}y{59pt}y{59pt}}
         & \multicolumn{4}{c}{Dataset: Warmstart (+ Expert); Sampling: Compositional}\\
         & Train Return & Test Return & Train Success & Test Success \\
         \hline
        Behavioral Cloning & $132.54 $ {\tiny $\pm 3.45\,\,\,$} & \,\,$51.04  $ {\tiny$\pm18.34\,\,\,$} & \shadecell{0.29} {\tiny $\pm0.01$} & \shadecell{0.07} {\tiny $\pm0.06$} \\
        Implicit Q-Learning & $\,\,\,\,98.64 $ {\tiny $\pm3.13$} & \,\,$57.82  $ {\tiny $\pm11.90$} & \shadecell{0.18} {\tiny $\pm 0.01$} & \shadecell{0.07} {\tiny $\pm0.03$} \\
        CP Behavioral Cloning & $153.36 $ {\tiny $\pm7.94\,\,\,$} & $89.86  $ {\tiny $\pm10.51$} & \shadecell{0.35} {\tiny $\pm0.01$} & \shadecell{0.17} {\tiny $\pm0.01$} \\
        CP Implicit Q-Learning & $127.75 $ {\tiny $\pm5.97\,\,\,$} & $87.31$ {\tiny $\pm22.72$} & \shadecell{0.30} {\tiny $\pm0.01$} & \shadecell{0.18} {\tiny $\pm0.10$} \\
        \end{tabular}
\end{table}

\begin{table}[t!]
    \caption{Test and training return and success rates achieved by Behavioral Cloning (BC), Implicit Q-Learning (IQL) and Compositional 
    BC (CP-BC) and Compositional IQL (CP-IQL) on the expert datasets in the restricted sampling setting. All agents  achieve decent training success. However, transfer to unseen tasks remains a challenge, especially for non-compositional agents. All values represent mean $\pm$ standard deviation.\\}
    \label{tab:holdoutres}
    \centering
        \begin{tabular}{y{110pt}|y{60pt}y{60pt}y{60pt}y{60pt}}
         & \multicolumn{4}{c}{Dataset: Expert; Fixed Element: \IIWA{}}\\
         & Train Return & Test Return & Train Success & Test Success \\
         \hline
        Behavioral Cloning & $371.57 $ {\tiny $\pm10.27$} & $18.54  $ {\tiny $\pm5.54$} & \shadecell{0.95}   {\tiny $\pm0.03$} & \shadecell{0.02}   {\tiny $\pm0.01$} \\
        Implicit Q-Learning & $273.71 $ {\tiny $\pm17.64$} & $34.85  $ {\tiny $\pm5.25$} & \shadecell{0.70}   {\tiny $\pm0.05$} & \shadecell{0.03}   {\tiny $\pm0.02$} \\
        CP Behavioral Cloning & $386.30 $ {\tiny $\pm \,\, \,4.79$} & $77.02  $ {\tiny $\pm40.17$} & \shadecell{0.98}   {\tiny $\pm0.02$} & \shadecell{0.11}   {\tiny $\pm0.11$} \\
        CP Implicit Q-Learning & $361.49 $ {\tiny $\pm \,\, \,8.82$} & $127.49  $ {\tiny $\pm25.98$} & \shadecell{0.95}   {\tiny $\pm0.02$} & \shadecell{0.18}   {\tiny $\pm0.04$} \\
        \end{tabular} \\
        \vspace{0.5em}
        \begin{tabular}{y{110pt}|y{60pt}y{60pt}y{60pt}y{60pt}}
         & \multicolumn{4}{c}{Dataset: Expert; Fixed Element: \pickPlace{}}\\
         & Train Return & Test Return & Train Success & Test Success \\
         \hline
        Behavioral Cloning & $346.84 $ {\tiny $\pm19.94$} & \,\,\,$41.76$ {\tiny $\pm \,\,\,9.27$} & \shadecell{0.88} {\tiny $\pm0.06$} & \shadecell{0.06} {\tiny $\pm0.03$} \\
        Implicit Q-Learning & $262.24 $ {\tiny $\pm13.35$} & \,\,\,$49.82  $ {\tiny $\pm11.46$} & \shadecell{0.64}   {\tiny $\pm0.05$} & \shadecell{0.07} {\tiny $\pm0.01$} \\
        CP Behavioral Cloning & $382.70$ {\tiny $\pm5.03$} & $81.83  $ {\tiny $\pm34.19$} & \shadecell{0.97} {\tiny $\pm0.01$} & \shadecell{0.13} {\tiny $\pm0.07$} \\
        CP Implicit Q-Learning & $368.92$ {\tiny $\pm7.38$} & $75.18  $ {\tiny $\pm21.34$} & \shadecell{0.93} {\tiny $\pm0.02$} & \shadecell{0.16} {\tiny $\pm0.08$} \\
        \end{tabular} \\
        \vspace{0.5em}
        \begin{tabular}{y{110pt}|y{60pt}y{60pt}y{60pt}y{60pt}}
         & \multicolumn{4}{c}{Dataset: Expert; Fixed Element: \hollowBox{}}\\
         & Train Return & Test Return & Train Success & Test Success \\
         \hline
        Behavioral Cloning & $363.83 $ {\tiny $\pm13.15$} & \,\,\,$45.38  $ {\tiny $\pm25.12$} & \shadecell{0.92} {\tiny $\pm0.03$} & \shadecell{0.08} {\tiny $\pm0.08$} \\
        Implicit Q-Learning & $278.81 $ {\tiny $\pm41.53$} & \,\,\,$63.48  $ {\tiny $\pm13.99$} & \shadecell{0.69} {\tiny $\pm0.12$} & \shadecell{0.11} {\tiny $\pm0.04$} \\ 
        CP Behavioral Cloning & $383.11 $ {\tiny $\pm0.62$} & $103.27  $ {\tiny $\pm \,\,\,24.06$} & \shadecell{0.97} {\tiny $\pm0.02$} & \shadecell{0.25} {\tiny $\pm0.05$} \\
        CP Implicit Q-Learning & $377.45 $ {\tiny $\pm0.69$} & $60.50 $ {\tiny $\pm \,\,\,4.32$} & \shadecell{0.97} {\tiny $\pm0.02$} & \shadecell{0.14} {\tiny $\pm0.01$} \\
        \end{tabular}\\
        \vspace{0.5em}
        \begin{tabular}{y{110pt}|y{60pt}y{60pt}y{60pt}y{60pt}}
         & \multicolumn{4}{c}{Dataset: Expert; Fixed Element: \objectWall}\\
         & Train Return & Test Return & Train Success & Test Success \\
         \hline
        Behavioral Cloning & $349.58 $ {\tiny $\pm14.81$} & $12.69  $ {\tiny $\pm\,\,\,3.31$} & \shadecell{0.88} {\tiny $\pm0.04$} & \shadecell{0.02} {\tiny $\pm0.03$} \\
        Implicit Q-Learning & $267.90 $ {\tiny $\pm20.08$} & $19.56  $ {\tiny $\pm\,\,\,2.39$} & \shadecell{0.64} {\tiny $\pm0.07$} & \shadecell{0.02} {\tiny $\pm0.02$} \\
        CP Behavioral Cloning & $393.10 $ {\tiny $\pm\,\,\,3.64$} & $41.42  $ {\tiny $\pm9.64$} & \shadecell{0.99} {\tiny $\pm0.01$} & \shadecell{0.10} {\tiny $\pm0.01$} \\
        CP Implicit Q-Learning & $377.43 $ {\tiny $\pm\,\,\,1.19$} & $23.39  $ {\tiny $\pm11.84$} & \shadecell{0.96} {\tiny $\pm0.01$} & \shadecell{0.04} {\tiny $\pm0.03$} \\
        \end{tabular}
\end{table}

\textbf{Training on uniformly sampled datasets}~~ To evaluate learnability and characterize different levels of challenge among our scenarios, we trained the BC, IQL, CP-BC and CP-IQL agents on the expert, medium, and medium-replay datasets. We used uniform sampling of $224$ training and $32$ zero-shot test tasks as discussed in section~\ref{sec:trainlists}. The results in the first three sub-tables of Table~\ref{tab:defaultres} verify that the four baselines can achieve high performance on the expert datasets. IQL baselines strictly outperform BC baselines in the settings where fewer successful trajectories are available (medium and replay), and generalize better to unseen configurations. When trained on replay data, BC attains no success, while IQL achieves some success. Further, while the compositional architecture boosts performance on the medium dataset, its generalization capabilities remain far from optimal.

\textbf{Training on Expert-Warmstart composition}~~ We demonstrate the importance of compositionality by evaluating agents using compositional sampling combining expert and warmstart datasets. As shown in the fourth sub-table of Table~\ref{tab:defaultres}, all four agents are able to extract some information from the expert datasets. 
The compositional architecture leads to an  increase in training performance, which translates to better zero-shot performance. This indicates that the learned modules discover how to solve pieces of tasks on the training set, which are then re-used on the zero-shot tasks.  
BC and standard IQL agents perform substantially better on the medium dataset (sub-table 2) even though the medium dataset and the warmstart-expert dataset with compositional sampling contain a similar amount of successful trajectories. This suggests that they learn something akin to an ``average'' policy, instead of extracting the compositional structure and specializing it to each task. 

\textbf{Training on restricted sampling}~~ As one additional test of compositionality, we compared agents on four restricted settings, each restricting one element from a distinct axis (Table~\ref{tab:holdoutres}). Agents were trained on expert data that only contains one task with the restricted element, while all zero-shot tasks contain the restricted element. The four agents perform well on the training tasks, but fail to generalize to the zero-shot tasks. Together with the uniform sampling results in Table~\ref{tab:defaultres}, these results demonstrate that the baselines require a large amount of data from varied task combinations for every single task element to generalize to unseen tasks. This is further evidence that current methods are incapable of extracting and leveraging the compositional structure of the environment. However, the compositional architecture shows signs of zero-shot generalization across all tasks, encouraging the study of compositional methods for robotic transfer learning problems.

\section{Related work} \label{sec:related}

\textbf{Compositional RL}~~Composition has been used in RL for decades in attempts to improve sampling efficiency~\citep{mendez2023how}. Intuitively, learning components of a problem may be easier than learning the full problem, and learned components can be combined with others to solve new tasks. The majority of such works in RL focused on learning temporally extended actions (skills or options) that can be sequenced to construct a compositional policy~\citep{sutton1999between,konidaris2009skilldiscovery,bacon2017option,tessler2017deep}. Other common forms of composition include logical composition~\citep{nangue2020boolean,barreto2018transfer,van2019composing}, state abstraction learning~\citep{dayan1993feudal,dietterich2000hierarchical,vezhnevets2017feudal}, and object-based RL~\citep{li2020learning,mu2020refactoring}. We consider the \textit{functional composition} perspective, described in Section~\ref{sec:preliminaries}, where components correspond to successive functional transformations of the state to generate actions~\citep{devin2017learning,goyal2021recurrent,mendez2022modular}. 

\textbf{Offline RL}~~ Offline RL research has grown steeply in recent years~\citep{LangeGabelRiedmiller2011chapter,levine2020offline}. Most methods operate in the single-task setting~\citep{fujimoto2019off,kumar2019bear,nair2020awac,kumar2020cql,ma2021conservative,kostrikov2022iql}, failing to leverage related datasets to train more powerful and general policies. Works on multi-task offline RL have been successful, but their scale remains limited~\citep{siegel2020keep, yu2021conservative}. Recently, large-scale datasets have enabled generalization of multi-task offline Q-learning~\citep{kumar2023offline}. Offline meta-RL, which pre-trains models on varied tasks to rapidly adapt to new tasks, shares motivation with our work. While most such methods consider offline fine-tuning~\citep{mitchell21offline,dorfman2021offline,li2021focal}, others instead adapt the policy to new tasks online via exploration~\citep{pong2022offline,zhao2022offline}. These prior works share information across tasks in an unstructured way, without considering common elements. Our explicitly compositional datasets promote the study of algorithms that reason about compositional relations across tasks. 

\textbf{Datasets and benchmarks}~~ Large image datasets have driven many advancements of deep learning~\citep{deng2009imagenet,krizhevsky2009learning}. Conversely, (online) RL has been restricted to the use of simulation benchmarks for assessing new methods' performance, primarily focused on single-task training~\citep{bellemare2013arcade,brockman2016openai,vinyals2017starcraft}. More recent work has developed online benchmarks for multi-task, lifelong, and meta-RL~\citep{james2020rlbench,cobbe2020leveraging,chevalierboisvert2018babyai,henderson2017multitask,ahmed2021causalworld,yu2020meta,tomilin2023coom}. With the advent of offline RL, it has become desirable to leverage large datasets for standardized RL training. Several such datasets have been proposed to benchmark offline RL~\citep{mandlekar2018roboturk, dasari2020robonet, fu2020d4rl, gulcehre2020rl, qin2022neorl, zhou2022train, liu2023datasets} and offline multi-agent RL~\citep{qu2023hokoff} approaches. However, none of these study the interplay of deep RL and compositionality. Our dataset construction follows D4RL~\citep{fu2020d4rl} to collect data from existing online benchmarks (in particular, CompoSuite; \citealp{mendez2022composuite}). Offline RL datasets are important for moving towards benchmarks on robot hardware~\citep{collins2020benchmarking, gurtler2023benchmarking}, and so our work relates to the development of low-cost~\citep{yang2019replab2, ahn2019robel} or remote-access~\citep{picked2017robotarium, paull2017duckietown, kumar2019offworld} robotic platforms for data collection.


\section{Limitations} \label{sec:limitations}

While our datasets are collected from simulation of commercially available robotic manipulators, it is well known that most learning algorithms suffer from a simulation-to-real (sim2real) performance gap. In consequence, work that seeks to apply learned policies or modules directly to physical robots would need to develop mechanisms to bridge this gap. Notably, compositionality might enable one such technique: training a new module for the physical robot in combination with pre-trained modules for the remaining task components. In addition, our focus is on releasing large-scale data sets to enable pre-training of powerful, generalizable offline RL models. As such, the computing requirements for running experiments on the full data sets would be prohibitive for organizations without access to powerful computers (see Appendix~\ref{sec:comp_req} for details of the computational setup used in our experiments). That being said, several of the proposed experimental settings require substantially less computation, making them more accessible for such organizations. Beside these limitations, our datasets inherit some of the limitations of the original CompoSuite benchmark. Namely, our datasets use symbolic (and not image) observations, the observation space reveals the compositional structure of the tasks explicitly, and the tasks contain a fixed number of four compositional axes~\citep{mendez2022composuite}. 

\section{Conclusion} \label{sec:conclusion}

In this paper we have introduced several novel datasets to study the intersection of offline and compositional RL. Our results indicate that current offline RL approaches do not capture the compositional structure of our tasks well, and that further research is required in this area. An interesting direction for future work is the explicit modeling of modularity in neural networks, or the discovery of modular structure, required to obtain networks that are capable of zero-shot generalization. Other directions include the study of offline to online transfer in a multi-task setting as well as a continual learning setting. An interesting open scientific question would be whether increasing the variety of compositional tasks has significant benefits over training on single tasks. We hope that, by releasing the datasets and the experimental settings described in this work, we can further research efforts in offline and compositional RL for robotics applications.

\subsubsection*{Acknowledgments}
\label{sec:ack}

JMM's research was funded by an MIT-IBM Distinguished Postdoctoral Fellowship. MH, AS, CK, and EE's research was partially supported by  DARPA Lifelong Learning Machines grant FA8750-18-2-0117, DARPA SAIL-ON contract HR001120C0040, DARPA ShELL agreement HR00112190133, Army Research Office MURI grant W911NF20-1-0080, and DARPA Triage Challenge award HR001123S0011. Any opinions, findings, and conclusion or recommendations expressed in this material are those of the authors and do not necessarily reflect the view of DARPA, the Army, or the US government.


\bibliography{main}
\bibliographystyle{rlc}

\appendix
\newpage

\section{Computational requirements} \label{sec:comp_req}

We ran our experiments using both server-grade (e.g., NVIDIA RTX A6000s) and consumer-grade (e.g., NVIDIA RTX 3090) GPUs, depending on the number of tasks we consider. Large experiment's training on $224$ tasks can be run within two days on a single NVIDIA A6000 GPU, but require up to $256$GB of RAM. Smaller experiments with up to $64$ training tasks can be trained within less than one day on a single RTX 3090 and $70$GB of RAM. For evaluation, we used consumer-grade AMD CPUs with $16$ cores and a single RTX 3090 for model inference.

\section{Hyperparameters}
\label{sec:hyperparameters}

With the exception of the batch size, hyperparameters were left at the default values used in d3rlpy. Table~\ref{tab:bc_hyper} contains the hyperparameters used to generate the BC results, Table~\ref{tab:iql_hyper} contains those for IQL. Compositional BC/IQL used the same hyperparamters as BC/IQL, with the exception of the neural network architecture, which is described in detail in Appendix~\ref{sec:DetailsCompositionalPolicy}. For the standard BC and IQL training, each neural network (all policies, Q-functions, and value functions) is encoded as a multi-layer perceptron (MLP) with $2$ hidden layers and $256$ hidden units per layer. 

\begin{table}[H]
    \centering
    \caption{Hyperparameters for Behavioral Cloning}
    \begin{tabular}{ | m{3.5cm} | m{2cm}| }
      \hline
      Optimizer & Adam \\ 
      \hline
      Adam $\beta_1$ & 0.9 \\ 
      \hline
      Adam $\beta_2$ & 0.999 \\ 
      \hline
      Adam $\varepsilon$ & 1e-8 \\ 
      \hline
      Learning Rate & 1e-3 \\ 
      \hline
      Batch Size
      & \#Tasks $\times 256$ \\ 
      \hline
    \end{tabular}
    \label{tab:bc_hyper}
\end{table}
\begin{table}[H]
    \hfill
    \centering
    \caption{Hyperparameters for Implicit Q-Learning}
    \begin{tabular}{ | m{3.5cm} | m{2cm}| }
      \hline
      Optimizer & Adam \\ 
      \hline
      Adam $\beta_1$ & $0.9$ \\ 
      \hline
      Adam $\beta_2$ & $0.999$ \\ 
      \hline
      Adam $\varepsilon$ & $1e-8$ \\ 
      \hline
      Actor Learning Rate & $4e-3$ \\ 
      \hline
      Critic Learning Rate & $4e-3$ \\ 
      \hline
      Batch Size
      & \#Tasks $\times 256$ \\ 
      \hline
      n\_steps & $1$ \\
      \hline
      $\gamma$ & $0.99$ \\
      \hline
      $\tau$ & $0.005$ \\
      \hline
      n\_critics & $2$ \\
      \hline
      expectile & $0.7$ \\
      \hline
      weight\_temp & $3.0$ \\
      \hline
      max\_weight & $100$ \\
      \hline
    \end{tabular}
    \label{tab:iql_hyper}
\end{table}
\renewcommand{\arraystretch}{0.9}
\section{Metrics}
\label{sec:metrics}
The metrics we report follow the original CompoSuite publication~\citep{mendez2022composuite}. The two metrics are given by:
\begin{itemize}
    \item per-task cumulative returns: $\overline{R} = \frac{1}{NM}\sum_{i=1}^{N}\sum_{j=1}^{M}\sum_{t=1}^{H} R_i(s_t,a_t)$, and
    \item per-task success rate: $\overline{S} = \frac{1}{NM} \sum_{i=1}^{N}\sum_{j=1}^{M} \max_{t\in[1,H]}  \mathds{1} [R_{i}(s_t, a_t) = 1]$,
\end{itemize}
where $N$ is the number of tasks, $M$ is the number of evaluation trajectories, the length of each trajectory is $H$, and  $\mathds{1}$ is the indicator function. A success is defined as reaching the maximum reward of $1$ per step in a single step during evaluation. Note that the success metric counts trajectories in which the agent is in a successful state at \textit{any} time. In consequence, if the agent receives the maximum step reward once but then moves to a non-successful configuration, the trajectory is still counted as successful. We evaluate these two metrics separately over the training tasks and the (remaining) zero-shot tasks.

\section{Details on Compositional Policy}
\label{sec:DetailsCompositionalPolicy}

Our compositional policies use the same neural network architecture as used by \citet{mendez2022composuite,mendez2022modular}, which follows a graph structure that exploits the compositional relations across CompoSuite tasks. The full network consists of $16$ MLP modules, each of which corresponds to a single element in CompoSuite---four obstacle modules, four object modules, four objective modules, and four robot modules. The graph is constructed hierarchically by passing the output of the previous module as (part of the) input to the next module. Each module operates in three stages: 1)~a pre-processing MLP that consumes the module-specific component of the state as input (e.g., the robot module processes only the proprioceptive state features), 2)~a concatenation layer that combines the output of the pre-processing module and the output of the previous module, and 3)~a post-processing MLP that consumes the concatenated input and produces the module's output. The order of the hierarchy is {\tt obstacle} $\rightarrow$ {\tt object} $\rightarrow$ {\tt objective} $\rightarrow$ {\tt robot}. The {\tt obstacle} modules have a single stage (since they are the first module), which is an MLP with a single hidden layer of size $32$. The MLPs of the {\tt object} module have each one hidden layer of $32$ units. The first stage of the {\tt objective} modules is an MLP with two hidden layers of $64$ units each, and the second stage is an MLP with a single hidden layer of size $64$. The {\tt robot} module's first stage MLP has three hidden layers of size $64$ and the second stage is the policy's output layer of dimension $8$ for the $8$ actions.

\end{document}